\documentclass{article}

\usepackage[preprint]{neurips_2026}

\usepackage[utf8]{inputenc} 
\usepackage[T1]{fontenc}    
\usepackage{hyperref}       
\usepackage{url}            
\usepackage{booktabs}       
\usepackage{amsfonts}       
\usepackage{nicefrac}       
\usepackage{microtype}      
\usepackage{xcolor}         
\usepackage{graphicx}
\usepackage{amsmath}
\usepackage{algorithm}
\usepackage{algorithmicx}
\usepackage[noend]{algpseudocode}
\usepackage{multirow}
\usepackage{caption}

\usepackage{tikz}
\usetikzlibrary{arrows.meta,positioning,fit,backgrounds,calc}

\title{From Regression to Inference: Meta-Learning Predictors for Neural Architecture Search}

\author{%
  Liping~Deng \\ 
  Department of Mathematics\\
  University of California, Riverside\\
  Riverside, CA 92507 USA\\
  \texttt{lipingd@ucr.edu} \\
  \And
  MingQing Xiao \\
  School of Mathematical and Statistical Sciences \\
  Southern Illinois University Carbondale\\
  Carbondale, IL 62901 USA\\
  \texttt{mxiao@siu.edu} \\
}

\begin{document}

\maketitle

\begin{abstract}
 Prediction-based approaches are widely used in neural architecture search (NAS), where a predictor estimates the performance of candidate architectures to guide selection. However, existing predictors are typically trained via supervised regression on limited samples, leading to overfitting and poor generalization to unseen architectures. In this work, we propose a fundamentally different formulation that models performance prediction as a {\it conditional function inference} problem using a Convolutional Neural Process (ConvNP) with meta-learning capabilities. Instead of fitting a fixed mapping to limited samples, our approach meta-learns to infer performance from partial observations by training with context–target splits across a group of synthesized tasks, explicitly optimizing for generalization under data scarcity and aligning the training procedure with the deployment setting in NAS. We further design simple yet effective meta-features for cell-based architectures and evaluate our method on NAS-Bench-101 and NAS-Bench-201. Extensive experiments show that our approach consistently improves top-K ranking quality and achieves the state-of-the-art architecture selection using limited samples.
 
\end{abstract}

\section{Introduction}
The performance of deep learning models is highly dependent on their architectures, motivating the development of neural architecture search (NAS) methods to automate architecture design  \cite{zoph2017neural,liu2018darts,white2023neural,ma2024single}. However, the vast size of modern search spaces makes exhaustive evaluation impractical, as training each candidate architecture can be computationally expensive. To address this challenge, prediction-based NAS has emerged as an efficient paradigm \cite{baker2017accelerating,liu2018progressive,lukasik2020neural,liu2022bridge}, where a learned predictor estimates the performance of candidate architectures and guides the search process. By avoiding full training of all candidates, such methods can significantly reduce computational cost.

Despite their success, existing predictors are typically trained via supervised regression, learning a fixed mapping from architecture representations to performance, e.g., random forest, multilayer perceptron, linear regression \cite{dudziak2020brp,pereira2023neural}, and graph neural networks (GCN)-based \cite{wen2020neural,ji2024cap2,ji2025carl}. In practice, however, NAS operates in a highly data-scarce regime, where only a tiny fraction of architectures can be observed. As a result, predictors must generalize from a small set of observed architectures to a much larger set of unseen candidates, often leading to overfitting and poor out-of-distribution performance. Moreover, conventional regression-based training does not explicitly reflect the deployment scenario in NAS, where predictions must be made conditioned on partial observations of the search space.

In this work, we address these limitations by reformulating performance prediction in NAS as a conditional function inference problem. Instead of learning a deterministic mapping, we model the predictor as a function conditioned on observed architecture–performance pairs. To this end, we adopt a Convolutional Neural Process (ConvNP \cite{foong2020meta}) that enables prediction of unseen architectures by learning from context–target splits across a distribution of tasks. To construct such tasks, we generate synthetic datasets via resampling and shuffling, allowing the model to meta-learn how to infer performance under varying observation patterns. This formulation explicitly aligns the training process with the inference scenario in NAS, improving generalization under limited data.

Furthermore, we design a simple yet effective set of meta-features to represent cell-based architectures and train the predictor using a combination of regression and ranking objectives, encouraging both accurate prediction and effective ordering of candidate architectures. We evaluate our approach on NAS-Bench-101 \cite{ying2019bench} and NAS-Bench-201 \cite{dong2020bench}, two widely used benchmarks for reproducible NAS. Extensive experiments demonstrate that ConvNP achieves competitive or superior predictive performance compared to existing methods, while consistently improving top-K architecture selection under low-data regimes.

Importantly, our results reveal a key insight: strong global ranking performance does not necessarily translate into better architecture selection. While some methods achieve higher rank correlation, they do not always identify the best-performing architectures. In contrast, our approach demonstrates that optimizing for selection-oriented metrics, such as top-K performance, leads to more effective NAS in practice. Our contributions are summarized as follows:
\begin{itemize}
    \item {\bf A new formulation of NAS prediction}. We reformulate architecture performance prediction as conditional function inference and propose a ConvNP-based predictor that models performance from partial observations instead of a fixed mapping.
    \item {\bf Meta-learning via synthetic task construction}. We propose a simple yet effective strategy for generating synthetic tasks and training with context–target splits, enabling the predictor to generalize under extreme data scarcity.
    \item {\bf A practical perspective on NAS evaluation}. We demonstrate that global ranking metrics (e.g., Kendall’s tau) may not align with the goal of architecture selection, and show that ConvNP consistently achieves strong top-K performance.
    \item {\bf Extensive empirical validation}. We conduct comprehensive experiments on NAS-Bench-101 and NAS-Bench-201, showing that ConvNP achieves competitive or superior results compared to state-of-the-art predictors, particularly in low-data regimes.
\end{itemize}

\section{Related work} \label{sec2}
\paragraph{Prediction-based NAS}

Prediction-based approaches estimate the performance of candidate architectures using a learned assessor, thereby avoiding expensive training during search. Early works encode architectures as sequences and apply neural or classical regression models \cite{deng2017peephole, liu2018darts,istrate2019tapas}, while more recent approaches adopt graph-based predictors \cite{wen2020neural,tang2020semi,shi2020bridging, ning2020generic,ji2025carl}, where architectures are represented as directed graphs and encoded using graph convolutional networks (GCNs). These methods either predict absolute performance or learn relative rankings, and semi-supervised variants further exploit relationships among architectures to propagate performance information.

In addition, some works rely on handcrafted meta-features, using statistical and structural properties of architectures with standard regression models \cite{white2020study,ning2022ta,pereira2023neural,akhauri2024encodings}. Others employ proxy tasks or synthetic benchmarks to approximate architecture quality at low cost \cite{li2021generic,liu2021homogeneous}. Despite differences in representation and modeling, these approaches share a common paradigm: learning a mapping from architecture characteristics to performance using a limited set of evaluated samples.

However, such predictors are fundamentally based on supervised learning and thus learn a fixed mapping from architectures to performance. Their effectiveness is therefore highly sensitive to data scarcity and distribution mismatch, and they often struggle to generalize beyond the observed architectures—a common scenario in NAS where only a small fraction of candidates can be evaluated.

\paragraph{Meta-learning in NAS}

Meta-learning has been explored in NAS to improve efficiency under limited data. Prior work focuses on transferring knowledge across tasks through learned architecture representations, weight initialization, or hypernetwork-based weight generation \cite{shaw2019meta,ha2017hypernetworks,brock2018smash}. One-shot and few-shot NAS methods \cite{elsken2020meta,zhao2021few} further exploit weight sharing or fast adaptation mechanisms, while recent approaches \cite{wang2022global} incorporate meta-learning into performance prediction or ranking to enable few-shot comparisons.

Nevertheless, these methods typically apply meta-learning at the level of architecture design or initialization, rather than within the predictor itself. When used for prediction, they are still combined with conventional regression or ranking models, which learn a fixed mapping and do not explicitly model inference from partial observations.


\section{Preliminaries} \label{sec3}

\subsection{Convolutional neural process}\label{subsec:np}
The neural processes (NPs) \cite{garnelo2018conditional,kim2019attentive} are meta-learning models designed for conditional function inference. Given a set of observed input–output pairs, e.g., $\mathcal{C}=\{(x^{(c)}, y^{(c)})\}^C_{c=1}$, referred to as the context set, and a set of query inputs, e.g., $\mathbf{x}_\mathcal{T}=\{x^{(t)}\}^T_{t=1}$, referred to as the target set, the NPs model the conditional distribution of target outputs given the context observations. Instead of learning a deterministic mapping, it learns a distribution over functions, enabling predictions that account for uncertainty and variability across tasks.

Especially, the convolutional NP (ConvNP) \cite{foong2020meta} is a convolutional neural network (CNN)-based model that consists of an encoder that aggregates information from the context set into latent representations, and a decoder that predicts target outputs conditioned on these representations: 
\begin{equation}
    p_\theta(\mathbf{y}_{\mathcal{T}}|\mathbf{x}_{\mathcal{T}};\mathcal{C})=\int p_\theta(\mathbf{y}_{\mathcal{T}}|\mathbf{x}_{\mathcal{T}};\mathbf{z})p_\theta(\mathbf{z}|\mathcal{C}){\rm d}\mathbf{z}, \label{def:cnp}
\end{equation}
where $\theta$ is the model parameters, and $\mathbf{z}$ is the sampled latent variables capturing global task information.

The training of ConvNP is performed across a distribution of tasks, where each task is randomly partitioned into context and target sets, enabling the model to learn to infer unseen outputs from partial observations. Often, it is required to maximize the maximum likelihood (ML) loss, i.e.,
\begin{equation}
    \mathcal{L}_{ML} = \log\Big(\frac{1}{L}\sum_{l=1}^{L}\prod^T_{t=1}p_\theta\big(y^{(t)}|x^{(t)},z_l\big)\Big), z_l\sim p_\theta(\mathbf{z}|\mathcal{C}), \label{loss:nll} 
\end{equation}
where $L$ represents the number of sampled latent variables in its learning process. Since ConvNP incorporates convolutional structures, which introduce translation equivariance and enable extrapolation beyond the support of observed inputs. This property is particularly desirable in NAS, where the predictor must generalize from a small subset of architectures to a much larger search space.

\subsection{Benchmark datasets}
We elaborate our approach on two widely used NAS benchmark datasets: NAS-Bench-101 \cite{ying2019bench} and NAS-Bench-201 \cite{dong2020bench}. Both benchmarks define cell-based search spaces, where each architecture is represented as a directed acyclic graph (DAG) describing the composition of operations and connections.

NAS-Bench-101 contains approximately 423k candidate architectures, where each cell comprises up to seven nodes and a limited set of operations, including $3\times3$ convolution, $1\times1$ convolution, and $3\times3$ max-pooling. NAS-Bench-201 provides a smaller but more structured search space with 15,123 architectures, evaluated across multiple datasets and training configurations. Each cell contains various operations chosen from zeroize, $3\times3$ convolution, $1\times1$ convolution, $3\times3$ average pooling, and skip connection. These benchmarks provide precomputed performance metrics for all architectures, enabling efficient and reproducible evaluation of NAS algorithms without retraining. 

\section{Methodology} \label{sec4}
Given a neural architecture search space with $M$ candidate architectures, we aim to predict the performance of the candidates using only a small subset of the evaluated samples. Let $\mathcal{D}=\{(\mathbf{x}_i, y_i)\}_{i=1}^N$ denote the observed architectures, where $\mathbf{x}_i\in\mathbb{R}^r$ represents the architecture encodings (meta-features) and $y_i$ denotes their ground-truth performance. The goal is to infer the performance of unseen architectures $\mathbf{x}_\mathcal{T}=\{\mathbf{x}_j\}_{j=1}^{M-N}$.

Unlike existing approaches that learn a deterministic mapping $y=f_{\theta}(\mathbf{x})$, we model this task as conditional function inference with a ConvNP:
\begin{equation}
    y=f_\theta(\mathbf{x};\mathcal{D}), \label{def:cnp2}
\end{equation}
where predictions are conditioned on the observed architecture–performance pairs. This formulation explicitly reflects the deployment scenario in NAS, where only a small subset of architectures is available, and the predictor must generalize to unseen candidates. We next show how to obtain the training task $\mathcal{D}$ and train the predictor $f_\theta$.

\subsection{Metadata extraction}
Suppose the observed architectures are $\{A_1, A_2,\cdots, A_N\}$. We next extract metadata, including meta-features $\mathbf{x}$ and the performance $y$, from each architecture. In general, evaluating the performance of a deep neural network is expensive following the standard training protocol. Thanks to the public benchmark datasets NAS-Bench-101 and NAS-Bench-201, we can efficiently obtain performance data without training.  
Denote the extracted performance of sampled architectures as $\{y_i\}_{i=1}^N$.

On the other hand, meta-features are a set of descriptive measures that characterize an architecture uniquely, and they should be extracted efficiently. Therefore, we propose a set of meta-features that focus on three aspects of an architecture, namely, statistical information, architectural complexity, and structural information. To simplify our discussion, here we focus on NAS-Bench-101.
\begin{itemize}
    \item \textbf{Statistical information:} The number of an operation in a cell, such as $3\times3$ Conv, $1\times1$ Conv, and $3\times3$ max-pooling, and the number of nodes.
    \item \textbf{Architectural complexity:} This group is calculated based on the graph of a cell. We first list all paths that start at the input node and end at the output node. Then, for each path, we assign weights to the operations it has based on their complexity and follow the principle: $3\times3$ Conv weights 3, $1\times1$ Conv weights 2, and max-pooling weights 1, since it is the least complex operation. At the end, we sum all weights in a path, and define four measures from all paths, i.e., the maximum and minimum weights, the mode, and the total weights of all paths. We also collect the number of trainable parameters of the complete model under a cell in this category, because it explicitly indicates the complexity of a model.
    \item \textbf{Structural information:} This category uses the adjacency matrix of a cell to describe its structure, and it comprises four measures: number of edges, ratio between the number of edges and the number of nodes, maximum number of outgoing edges of a node, and maximum number of incoming edges of a node.
\end{itemize}

For NAS-Bench-201, due to the fixed graph structure in the search space, we ignore the number of nodes and structural measures since they are the same for all candidate architectures. The weights assigned to the five operations $3\times3$ Conv, $1\times1$ Conv, $3\times3$ average-pooling, skip connection, and zeroize are 5, 4, 3, 2, and 1, respectively. It is worth noting that even if the proposed meta-features are designed for the two benchmark search spaces, they can be generalized to other cell-based search spaces easily. Finally, let denote the meta-features of the sample architectures as $\{\mathbf{x}_i\}_{i=1}^N$, where each $\mathbf{x}_i$ is a vector of 13 (NAS-Bench-101) or 10 (NAS-Bench-201) measures. Finally, each measure is normalized to $[0, 1]$ using min-max scaling over the search space to eliminate inconsistent magnitudes.

\subsection{Training}
Given the context dataset $\mathcal{D}=\{(\mathbf{x}_i,y_i)\}_{i=1}^N$ of the observed architectures, where $N$ is often small, we show how to train an effective ConvNP predictor with limited training samples via meta-learning.

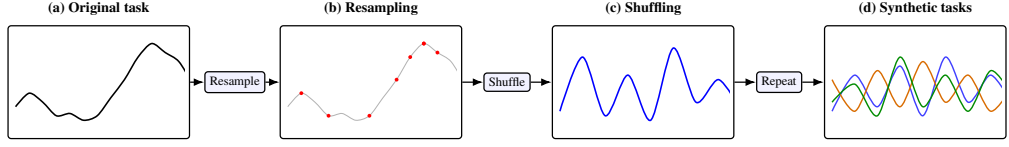
\begin{figure*}[t]
\centering
\resizebox{0.95\textwidth}{!}{%
\begin{tikzpicture}[
    >=Latex,
    font=\small,
    panel/.style={draw, rounded corners=2pt, thick, inner sep=0pt, fill=white},
    boxlabel/.style={font=\bfseries\small},
    op/.style={draw, rounded corners=2pt, thick, fill=blue!5, inner sep=3pt, align=center, font=\footnotesize},
    arrow/.style={->, thick},
    sampled/.style={red, fill=red},
    synth1/.style={blue!75, line width=0.9pt},
    synth2/.style={orange!85!black, line width=0.9pt},
    synth3/.style={green!55!black, line width=0.9pt}
]

\def\pw{4.0cm}
\def\ph{2.5cm}

\node[panel, minimum width=\pw, minimum height=\ph] (orig) at (0,0) {};
\node[boxlabel, anchor=south] at ($(orig.north)+(0,0.1)$) {(a) Original task};

\node[panel, minimum width=\pw, minimum height=\ph] (resamp) at (6,0) {};
\node[boxlabel, anchor=south] at ($(resamp.north)+(0,0.1)$) {(b) Resampling};

\node[panel, minimum width=\pw, minimum height=\ph] (shuffle) at (12,0) {};
\node[boxlabel, anchor=south] at ($(shuffle.north)+(0,0.1)$) {(c) Shuffling};

\node[panel, minimum width=\pw, minimum height=\ph] (synth) at (18,0) {};
\node[boxlabel, anchor=south] at ($(synth.north)+(0,0.1)$) {(d) Synthetic tasks};

\node[op] (op1) at (3,0) {Resample};
\node[op] (op2) at (9,0) {Shuffle};
\node[op] (op3) at (15,0) {Repeat};

\draw[arrow] (orig.east) -- (op1.west);
\draw[arrow] (op1.east) -- (resamp.west);

\draw[arrow] (resamp.east) -- (op2.west);
\draw[arrow] (op2.east) -- (shuffle.west);

\draw[arrow] (shuffle.east) -- (op3.west);
\draw[arrow] (op3.east) -- (synth.west);

\begin{scope}
\clip ($(orig.south west)+(0.12,0.12)$) rectangle ($(orig.north east)+(-0.12,-0.12)$);
\begin{scope}[shift={($(orig.south west)+(0.18,0.22)$)}]
\draw[black, line width=1pt]
plot[smooth] coordinates {
(0,0.5) (0.3,0.8) (0.6,0.6) (0.9,0.3) (1.2,0.35)
(1.5,0.2) (1.8,0.3) (2.1,0.7) (2.4,1.1) (2.7,1.6)
(3.0,1.9) (3.3,1.7) (3.6,1.5) (3.9,1.0) (4.2,0.4)
};
\end{scope}
\end{scope}

\begin{scope}
\clip ($(resamp.south west)+(0.12,0.12)$) rectangle ($(resamp.north east)+(-0.12,-0.12)$);
\begin{scope}[shift={($(resamp.south west)+(0.18,0.22)$)}]
\draw[gray!60]
plot[smooth] coordinates {
(0,0.5) (0.3,0.8) (0.6,0.6) (0.9,0.3) (1.2,0.35)
(1.5,0.2) (1.8,0.3) (2.1,0.7) (2.4,1.1) (2.7,1.6)
(3.0,1.9) (3.3,1.7) (3.6,1.5) (3.9,1.0) (4.2,0.4)
};
\foreach \x/\y in {
0.3/0.8, 0.9/0.3, 1.8/0.3,
2.4/1.1, 2.7/1.6, 3.0/1.9,
3.3/1.7, 3.9/1.0, 4.2/0.4,
3.0/1.9
}{
\fill[red] (\x,\y) circle (1.2pt);
}
\end{scope}
\end{scope}

\begin{scope}
\clip ($(shuffle.south west)+(0.12,0.12)$) rectangle ($(shuffle.north east)+(-0.12,-0.12)$);
\begin{scope}[shift={($(shuffle.south west)+(0.18,0.22)$)}]
\draw[blue, line width=1pt]
plot[smooth] coordinates {
(0,0.4) (0.5,1.6) (1.0,0.3) (1.5,1.2)
(2.0,0.2) (2.5,1.8) (3.0,0.6) (3.5,1.1)
(4.0,0.4)
};
\end{scope}
\end{scope}

\begin{scope}
\clip ($(synth.south west)+(0.12,0.12)$) rectangle ($(synth.north east)+(-0.12,-0.12)$);
\begin{scope}[shift={($(synth.south west)+(0.18,0.22)$)}]

\draw[synth1] plot[smooth] coordinates {
(0,0.4) (0.5,1.2) (1.0,0.5) (1.5,1.4)
(2.0,0.3) (2.5,1.6) (3.0,0.6) (3.5,1.2) (4.0,0.5)
};

\draw[synth2] plot[smooth] coordinates {
(0,1.1) (0.5,0.4) (1.0,1.3) (1.5,0.5)
(2.0,1.5) (2.5,0.6) (3.0,1.2) (3.5,0.4) (4.0,1.0)
};

\draw[synth3] plot[smooth] coordinates {
(0,0.6) (0.5,1.0) (1.0,0.3) (1.5,1.6)
(2.0,0.5) (2.5,1.2) (3.0,0.4) (3.5,1.3) (4.0,0.7)
};

\end{scope}
\end{scope}

\end{tikzpicture}
}
\caption{Synthetic task generation for ConvNP. A single observed task is transformed into multiple training tasks via resampling with replacement and random shuffling.}
\label{fig:synth}
\end{figure*}

\paragraph{Synthetic task generation} To meta-train the model under the formulation (\ref{def:cnp2}), we need a distribution of tasks. Since only a single dataset $\mathcal{D}$ is available, we generate $S$ synthetic tasks via resampling with replacement. Each task consists of a subset of architecture–performance pairs with a fixed number of points, with random ordering to encourage permutation invariance. Despite its simplicity, this data augmentation strategy produces a large and diverse set of tasks with varied behaviors, see Figure \ref{fig:synth}, where we assume the task has 1D inputs for illustration.


The standard training protocol of convolutional neural networks (CNNs) is adopted here. During a training epoch, a batch of tasks is randomly taken from the $S$ synthetic tasks. Then, each task is split into a context set $\mathcal{C}$ and a target set $\mathcal{T}$ exclusively and exhaustively. Note that the splitting in each batch should be different to encourage the model to learn from various scenarios. This process simulates the inference scenario in NAS, where predictions must be made based on partial observations of the search space.


\paragraph{Learning objective}
Given a synthetic task, let the context set be $\mathcal{C}=\{(\mathbf{x}^{(c)},y^{(c)})\}_{c=1}^C$ and the target set be $\mathcal{T}=\{(\mathbf{x}^{(t)},y^{(t)})\}_{t=1}^T$. Conditioned on the context set, the ConvNP predictor outputs the predictive means for each target architecture,
\begin{equation}
    \{\hat{y}_l^{(t)}\}_{l=1}^L = f_\theta(\mathbf{x}^{(t)};\mathcal{C}),\  t=1,2,\cdots, T. \label{def:pred}
\end{equation}
Recall that $L$ is the number of sampled latent variables; that is, the ConvNP generates $L$ predictions for each architecture.

Since the goal of NAS is to identify high-performing architectures rather than predicting distributions, instead of using the loss (\ref{loss:nll}), we train the model with a combination of regression and ranking objectives. The regression term encourages accurate prediction of validation accuracy, while the ranking term enforces correct relative ordering among candidate architectures in the target set. Specifically, the regression loss is defined as
\begin{equation}
    \mathcal{L}_{\rm MSE} = \frac{1}{LT}\sum_{l=1}^L\sum_{t=1}^T(y^{(t)}-\hat{y}_l^{(t)})^2. \label{loss:mse}
\end{equation}
For the ranking objective, we consider pairs of target architectures and encourage the predicted ordering to match the ground-truth ordering. Let $\mathcal{P}=\{(i,j):y^{(i)}>y^{(j)}\}$ be the set of ordered target pairs. We adopt a pairwise margin-based ranking loss:
\begin{equation}
    \mathcal{L}_{\rm rank} = \frac{1}{L|\mathcal{P}|}\sum_{l=1}^L\sum_{(i,j)\in\mathcal{P}}\max(0, m-(\hat{y}_l^{(i)}-\hat{y}_l^{(j)})), \label{loss:rank}
\end{equation}
where $m>0$ is a margin hyperparameter, and $|\mathcal{P}|$ stands for the cardinality of set $\mathcal{P}$.

Therefore, the overall training objective is
\begin{equation}
    \mathcal{L}_{\rm total} = \mathcal{L}_{\rm MSE}+\lambda\mathcal{L}_{\rm rank}, \label{loss:hybrid}
\end{equation}
where $\lambda>0$ balances prediction accuracy and ranking consistency. This objective aligns the learning process with the deployment goal in NAS: selecting architectures with strong relative performance under limited observations.

Although ConvNP naturally supports probabilistic prediction, in our setting, we focus on the predictive mean and optimize it directly with regression and ranking supervision. This is because architecture selection in NAS primarily depends on relative ordering, and empirically, we found likelihood-based objectives, e.g., Eq. (\ref{loss:nll}), to be less effective and less efficient than mean prediction. 


\begin{algorithm}
\caption{\texttt{ConvNP-based NAS}}\label{alg:convnp}
\hspace*{0.02in} {\bf Input:} Search space $\{A_1,\cdots,A_M\}$, number of sampled architectures $N$, number of synthetic tasks $S$, number of evaluated architectures $K$\\
\hspace*{0.02in} {\bf Output:} The recommended architecture $A^\star$
\begin{algorithmic}[1]
\Statex {\textbf{Training phase:}}
	\State {Randomly sample $N$ architectures $\{A_1,\cdots,A_N\}$}
	\State {Extract the context dataset $\mathcal{D}=\{\mathbf{x}_i,y_i\}_{i=1}^N$}
    \State {Generate $S$ synthetic tasks by resampling $\mathcal{D}$}
    \State {Train $f_\theta$ using the $S$ synthetic tasks by minimizing $\mathcal{L}_{\rm total}$}
\Statex	{\textbf{Prediction:}}
    \State {Extract meta-features from remaining $M-N$ architectures: $\mathbf{x}_\mathcal{T}=\{\mathbf{x}_{N+1},\cdots,\mathbf{x}_M\}$}
    \State {Make predictions: $\{\hat{\mathbf{y}}_l\}_{l=1}^L=f_\theta(\mathbf{x}_\mathcal{T};\mathcal{D})$, where $\mathbf{y}_l\in\mathbb{R}^{M-N}$}
    \State {Choose the top-K architectures based on $\frac{1}{L}\sum_{l=1}^L{\rm rank}(\hat{y}_l)$}
    \State {Evaluate the $K$ architectures and determine $A^\star$}
\end{algorithmic}
\end{algorithm}

\subsection{NAS via ConvNP}

After training, the predictor is applied to the rest of the search space. The observed architectures $\mathcal{D}$ serve as the context set, and predictions are made for unseen candidates. We select the top-K architectures based on predicted performance and evaluate them to identify the final solution. This results in a total evaluation cost of $N+K$, significantly reducing the search budget. The whole process of our approach is summarized in \texttt{Algorithm} \ref{alg:convnp}. In Step 6, since there are $L$ predictions, we first assign ranks to each prediction $\mathbf{y}_l$ using the standard ranking scheme, then the average rank across the $L$ ranking lists is calculated to form a final ranking list, from which we choose the top-K architectures in Step 7.


\section{Experiments} \label{sec5}

The experiments mainly comprise two parts: comparisons with various baseline predictors and ablation studies. We next introduce the adopted benchmark datasets, ConvNP model, and evaluation metrics before discussing the empirical results.

\subsection{Experimental settings}
\paragraph{Benchmark datasets}
We evaluate ConvNP on NAS-Bench-101 (NAS101) \cite{ying2019bench} and NAS-Bench-201 (NAS201) \cite{dong2020bench}. NAS101 contains $\sim$423K architectures evaluated on CIFAR-10, while NAS201 includes $\sim$15K architectures evaluated on CIFAR-10, CIFAR-100, and ImageNet16-120. Following prior work, we use validation and test accuracies at epoch 108 for NAS101 and epoch 200 for NAS201.

\paragraph{ConvNP model}
We adopt the architecture in \cite{dubois2020npf} with a minor modification: a two-layer MLP is added to map meta-feature vectors to scalar inputs, matching the encoder design. The model is trained using Adam with a learning rate of 0.001 and a batch size of 16. The number of latent variables $L\in\{1,4,8,16,32\}$ is selected per dataset based on validation performance. The $\lambda$ in loss (\ref{loss:hybrid}) is set to be 1000, and the parameter $m$ in the ranking loss (\ref{loss:rank}) is fixed as 0.01. All experiments are conducted on a single NVIDIA P100 GPU (16GB).

\paragraph{Evaluation metrics}
We report the validation and test accuracies of the best architecture selected from the top-30 predictions. We also evaluate ranking quality using Kendall’s tau and Recall@k. Kendall’s tau measures global ranking consistency, while Recall@k is a standard top-K evaluation metric in information retrieval \cite{schutze2008introduction}, measuring the proportion of relevant items retrieved among the top-K predictions. Since NAS primarily aims to identify high-performing candidates, we emphasize top-K performance over full ranking accuracy.

\subsection{Comparisons with baseline predictors}
In this experiment, we randomly sample 172 and 90 architectures from NAS101 and NAS201, respectively, and use them to generate 10,000 synthetic tasks with lengths of 86 (NAS101) and 90 (NAS201) for training the ConvNP predictor. The model is trained for one epoch. In each training batch, up to 10 samples from a synthetic task are randomly selected as the context set, while the entire task is used as the target set.

We compare ConvNP against five regression-based predictors, namely, linear regression (LR), ridge regression (RR), multilayer perceptron (MLP), random forest (RF), and eXtreme Gradient Boosting (XGB) \cite{chen2016xgboost}; three GCN-based predictors, including neural predictor (NP) \cite{wen2020neural}, context-aware predictor (CAP) \cite{ji2024cap2}, and causality-guided architecture representation learning (CARL) \cite{ji2025carl}, as well as two search algorithms: random search (RS) and regularized evolution (RE) \cite{real2019regularized}. For fair comparison, all regression models use the same meta-features as proposed in this work, and their hyperparameters are tuned via 5-fold cross-validation with 30 random search trials. The hyperparameters of the GCN-based predictors are set according to their official implementations.

\begin{figure}
\centering
\begin{minipage}[t]{0.45\textwidth}
    \vspace{0pt}
    \centering
    \footnotesize
    \begin{tabular}{lccc}
        \hline
        &Validation &Test &Ktau  \\
        \hline
        LR	&$94.37\pm0.14$		&$93.80\pm0.20$		&0.46	\\
        RR	&$94.47\pm0.20$		&$93.88\pm0.16$		&0.47	\\
        MLP	&$94.05\pm0.72$		&$93.47\pm0.67$		&0.31	\\
        RF	&$94.31\pm0.19$		&$93.71\pm0.20$		&0.53	\\
        XGB	&$94.13\pm0.34$		&$93.59\pm0.35$		&0.49	\\
        NP	&$94.64\pm0.27$		&$93.74\pm0.26$		&0.62	\\
        CAP	&$94.27\pm0.74$		&$93.50\pm0.70$		&0.59	\\
        CARL&$94.71\pm0.21$		&$93.93\pm0.19$		&0.68\\	
        \hline
        RS	&$94.15\pm0.11$ 	&$93.55\pm0.12$		&$-$	\\
        RE	&$94.37\pm0.21$		&$93.68\pm0.22$		&$-$	\\
        \hline
        {\bf ConvNP}	&$94.76\pm0.19$		&$94.09\pm0.12$		&0.57	\\
        Best&$95.06$		&$94.32$		&$-$	\\
        \hline
    \end{tabular}
    \captionof{table}{The comparisons on validation/test accuracies and Kentall's tau over NAS101.}
    \label{Tb:nas101}
\end{minipage}
\hspace{0.4in}
\begin{minipage}[t]{0.46\textwidth}
    \vspace{0pt}
    \centering
    \includegraphics[width=0.9\linewidth]{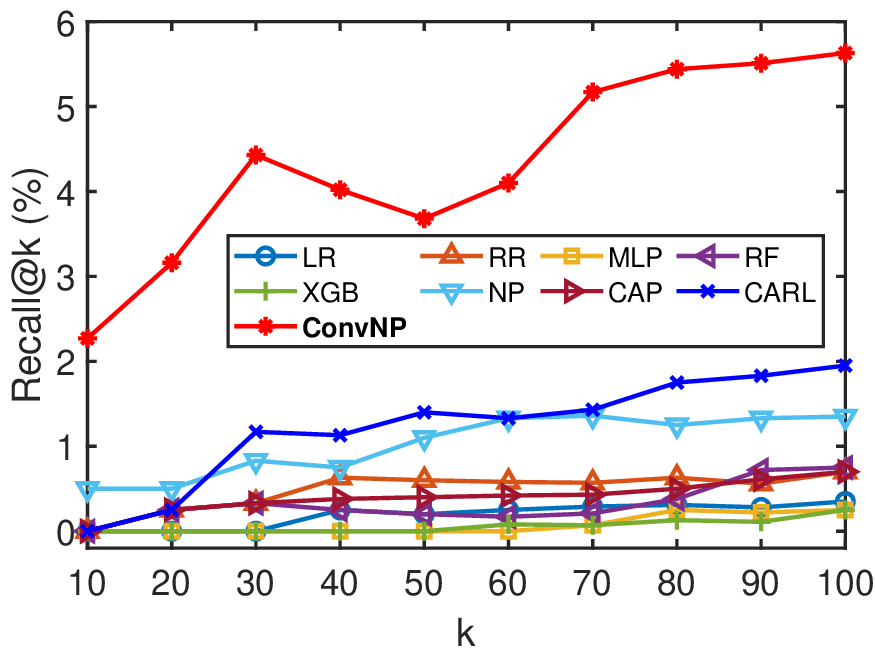}
    \caption{The comparisons on Recall@k over NAS101, where $k$ ranges from 10 to 100.}
    \label{fig:nas101-recall}
\end{minipage}
\end{figure}

\begin{table*}
\centering
\caption{The comparisons on validation and test accuracies over NAS201.} \label{Tb:nas201}
\footnotesize
\begin{tabular}{lcccccc}
\toprule
\multirow{2}{*}{} &
\multicolumn{2}{c}{CIFAR-10} &\multicolumn{2}{c}{CIFAR-100} & \multicolumn{2}{c}{ImageNet16} \\
\cline{2-7}
&Validation &Test &Validation &Test &validation &Test\\
\hline
LR	&   $90.72\pm0.51$ 	&	$93.84\pm0.31$ 	&	$71.51\pm0.97$ 	&	$71.84\pm0.92$ 	&   $45.07\pm1.19$  &	$45.59\pm1.01$ 		\\	
RR	&   $91.23\pm0.58$	&   $94.18\pm0.33$	&   $72.54\pm1.08$  &	$72.77\pm0.87$ 	&   $45.78\pm0.50$  &   $46.12\pm0.32$	\\
MLP &	$90.70\pm0.67$	&	$93.63\pm0.49$	&	$71.78\pm1.08$ 	&	$71.94\pm0.97$ 	&   $45.42\pm1.14$  &   $45.76\pm1.22$	\\
RF	&	$91.19\pm0.45$ 	&	$94.05\pm0.27$	&	$72.15\pm0.95$ 	&   $72.34\pm0.82$  &   $46.15\pm0.54$  &	$46.39\pm0.59$\\
XGB	&	$90.21\pm1.14$ 	&	$93.08\pm1.32$ 	&	$71.09\pm2.93$	&	$71.06\pm2.87$ 	&   $45.73\pm0.87$  &	$45.94\pm0.79$ 		\\	
NP	&   $90.87\pm0.48$	&   $93.92\pm0.38$	&   $71.48\pm1.43$  &	$71.84\pm1.37$ 	&   $46.19\pm0.17$  &   $46.41\pm0.09$	\\
CAP &	$90.91\pm0.63$ 	&	$93.81\pm0.51$ 	&	$72.24\pm1.18$ 	&	$72.32\pm1.18$ 	&   $46.00\pm0.57$  &	$46.49\pm0.60$\\
CARL&	$91.49\pm0.24$ 	&	$94.31\pm0.14$ 	&	$72.84\pm0.81$ 	&   $72.92\pm0.72$  &   $46.27\pm0.27$  &   $46.61\pm0.38$	\\
\hline
RS &	$91.02\pm0.22$ 	&	$93.83\pm0.24$ 	&	$71.42\pm0.76$ 	&	$71.55\pm0.81$ 	&   $45.40\pm0.65$  &   $45.80\pm0.65$	\\
RE &	$91.26\pm0.24$ 	&	$93.88\pm0.31$ 	&	$72.35\pm0.83$ 	&  $72.13\pm0.99$ &   $46.01\pm0.62$  &	  $45.79\pm0.95$\\
\hline
{\bf ConvNP} &	$91.53\pm0.14$ 	&	$94.31\pm0.12$ 	 &	$72.94\pm0.88$ 	&	$72.99\pm0.71$ 	&   $46.24\pm0.62$ &    $46.51\pm0.52$	\\
Best&	91.61 	&	94.37 	&	73.49 	&73.51   &46.73    &47.31	\\
\bottomrule
\end{tabular}
\end{table*}

\paragraph{Results on NAS101}
The empirical results are shown in Table \ref{Tb:nas101} and Figure \ref{fig:nas101-recall}, where all methods use the same training samples, generated across 20 random seeds. The results in Table 1 show that the proposed method achieves strong predictive performance, with a validation accuracy of 94.76\% and a test accuracy of 94.09\%, outperforming most baselines, including the search algorithms RS and RE, which start from 172 sampled architectures and perform 30 search iterations. In addition, ConvNP attains a Kendall’s tau of 0.57, outperforming traditional regression models such as LR, RR, MLP, RF, and XGB by a clear margin, indicating a better ability to preserve the relative ranking of architectures. While it does not surpass GCN-based methods (NP, CAP, and CARL), it still achieves comparable accuracy and demonstrates a solid balance between regression fidelity and ranking consistency.

Figure \ref{fig:nas101-recall} further highlights the effectiveness of the method from a decision-making perspective. Although its Kendall’s tau is not the highest, the Recall@k curve shows that it consistently identifies high-performing architectures, especially as $k$ increases. This indicates that the model is particularly effective for top-K selection, which is often more critical in NAS scenarios than global ranking accuracy. In contrast, methods such as CARL achieve higher overall rank correlation but do not consistently translate this advantage into better top-K recall. 

\paragraph{Results on NAS201}
From Table \ref{Tb:nas201}, ConvNP remains competitive in terms of predictive accuracy, achieving the near-best test accuracy on CIFAR-10 (94.31\%) and strong performance on CIFAR-100 (72.99\%) and ImageNet16 (46.51\%). Among the five traditional regressors, RR and RF perform relatively better, although their results are less stable. Among the three GCN-based predictors, CARL shows strong performance; it slightly outperforms ConvNP on ImageNet16 with a test accuracy of 46.61\% and achieves the highest Kendall’s tau across all three datasets, as shown in Table~\ref{Tb:nas201-ktau}. In contrast, ConvNP consistently outperforms the five regression models, as well as NP and CAP. 

\begin{figure}
    \centering
    \includegraphics[width=0.325\linewidth]{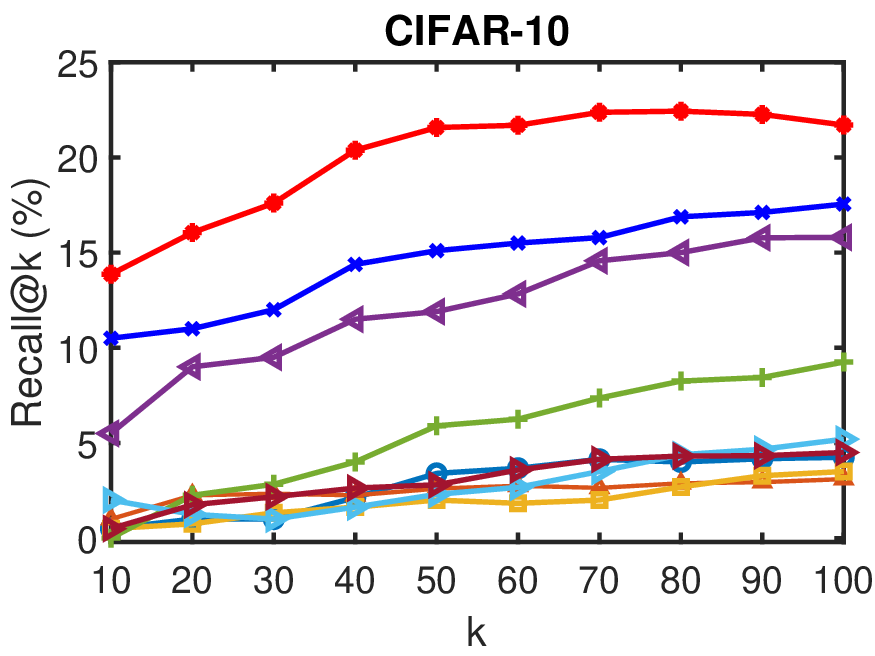}\
    \includegraphics[width=0.325\linewidth]{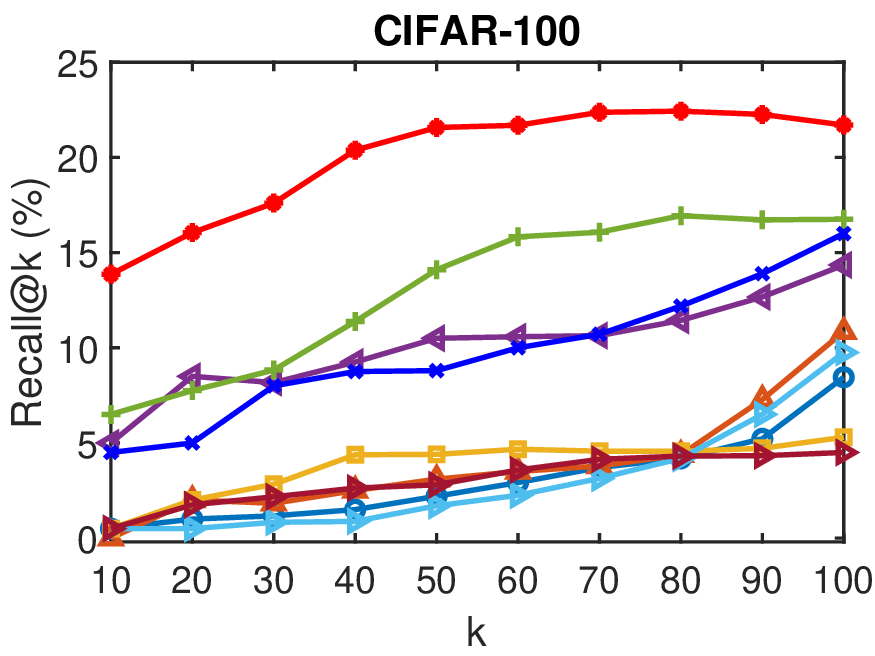}\
    \includegraphics[width=0.325\linewidth]{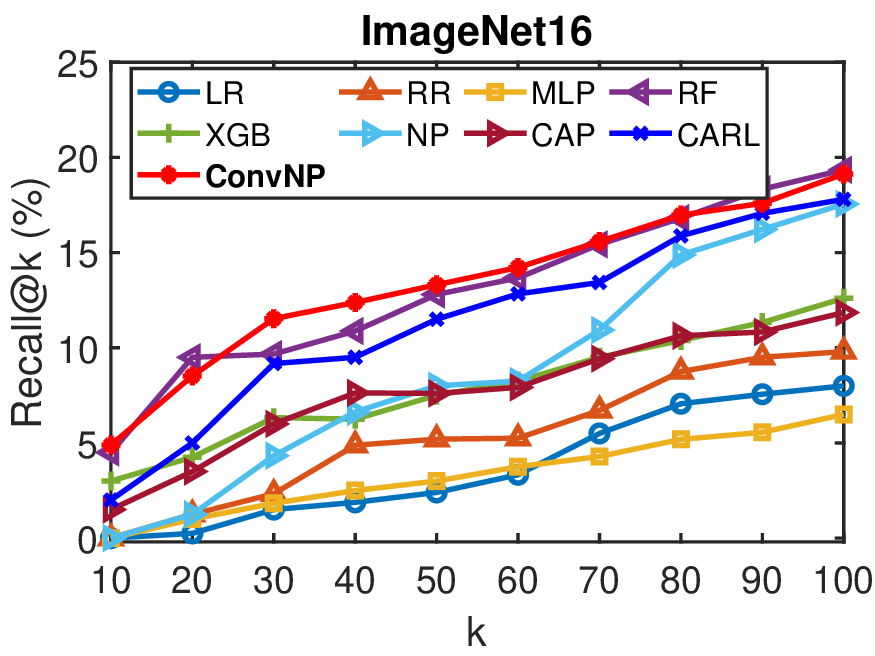}
    \caption{The comparisons on Recall@k over NAS201.}
    \label{fig:nas201-recall}
\end{figure}

Figure \ref{fig:nas201-recall} again reveals an important strength of ConvNP from a practical NAS perspective. Its Recall@k performance is consistently strong across all three datasets, particularly on CIFAR-10 and CIFAR-100, where it ranks among the top methods as $k$ increases. This reemphasizes a key observation: higher rank correlation (e.g., CARL) does not necessarily translate directly into superior top-K selection. 

\begin{table*}
\centering
\caption{The comparisons on Kendall's tau over NAS201.} \label{Tb:nas201-ktau}
\small
\begin{tabular}{lccccccccc}
\toprule
Benchmarks  &LR    &RR    &MLP   &RF    &XGB   &NP    &CAP   &CARL  &{\bf ConvNP}\\
\hline
CIFAR-10    &0.53  &0.57  &0.37  &0.49  &0.45  &0.47  &0.46  &0.64  &0.59	\\
CIFAR-100   &0.57  &0.60  &0.47  &0.55  &0.52  &0.59  &0.54  &0.65  &0.61	\\
ImageNet16  &0.58  &0.59  &0.48  &0.56  &0.55  &0.56  &0.57  &0.63  &0.59	\\
\bottomrule
\end{tabular}
\end{table*}

Overall, the comparisons over both NAS101 and NAS201 indicate that ConvNP can achieve a favorable trade-off between regression accuracy and practical selection performance, even though it is less competitive in pure global ranking metrics. The relatively lower Kendall’s tau of ConvNP may stem from latent variable sampling in its learning process, which introduces stochasticity into the predictions. Addressing this limitation is a promising direction for future work.

\subsection{Performance under varying training set sizes}
In this experiment, we examine the behavior of our method across varying training set sizes, specifically $ 50, 90, 172, 344, 516, 688$, and $860$ architectures. On NAS101, these correspond to approximately 0.012\% to 0.2\% of the full search space, covering an extreme-to-moderate low-data regime, whereas on NAS201, they represent about 0.3\% to 5.5\%, ranging from low to relatively large data settings. The empirical results are summarized in Table~\ref{Tb:nas101-201}. For brevity, we report results on CIFAR-10 (NAS101) and ImageNet16 (NAS201) only. 

The sampled architectures are constructed in a nested manner: we first draw 860 architectures uniformly at random, then iteratively form smaller subsets (688, 516, etc.) from this pool. This design allows us to consistently evaluate the effect of increasing training data. For each setting, we generate 10K synthetic tasks with sequence length 86, except for the smallest cases (50 and 90), where the task lengths match the dataset size. All other ConvNP configurations remain the same as in the previous experiments.

\begin{table*}
\centering
\caption{The performance of the ConvNP predictor under various training set sizes.} \label{Tb:nas101-201}
\footnotesize
\begin{tabular}{lcccccccc}
\toprule
\multirow{2}{*}{} &
\multicolumn{4}{c}{CIFAR-10 (NAS101)} &\multicolumn{4}{c}{ImageNet16 (NAS201)} \\
\cline{2-9}
&Validation &Test &Ktau &R@100 &Validation &Test &Ktau &R@100\\
\hline
50	&   $94.06\pm0.16$ 	&	$93.89\pm0.15$ 	&	0.48 	&2.08&	$45.82\pm0.90$ 	&   $46.16\pm0.72$  &	0.55  &13.67		\\	
90	&   $94.56\pm0.18$	&   $93.93\pm0.15$	&   0.52    &3.56&	$46.24\pm0.62$ 	&   $46.51\pm0.52$  &   0.59  &19.13	\\
172 &	$94.76\pm0.19$	&	$94.09\pm0.12$	&	0.57 	&5.63&	$46.21\pm0.35$ 	&   $46.53\pm0.25$  &   0.61  &20.52	\\
344	&	$94.72\pm0.21$ 	&	$94.07\pm0.13$	&	0.58 	&6.98&  $46.28\pm0.33$  &   $46.58\pm0.32$  &	0.62  &21.29 \\
516	&	$94.71\pm0.17$ 	&	$94.07\pm0.09$ 	&	0.59	&8.18&	$46.27\pm0.21$ 	&   $46.50\pm0.10$  &	0.62  &21.46		\\	
688	&   $94.70\pm0.20$	&   $94.06\pm0.12$	&   0.59    &7.73&	$46.24\pm0.24$ 	&   $46.47\pm0.16$  &   0.63  &20.57	\\
860 &	$94.74\pm0.23$ 	&	$94.08\pm0.14$ 	&	0.59 	&7.74&	$46.30\pm0.28$ 	&   $46.55\pm0.20$  &	0.63  &21.83\\
\bottomrule
\end{tabular}
\end{table*}


The results show that increasing the number of training samples from 50 to 860 consistently improves ConvNP's performance on both NAS101 and NAS201, with the largest gains in the low-data regime. On CIFAR-10 (NAS101), while validation and test accuracies quickly saturate around 94.7\% and 94.07\%, Kendall’s tau improves substantially from 0.48 to 0.59, and Recall@100 increases from 2.08 to a peak of 8.18, indicating a strong enhancement in ranking quality and top-K selection. Similarly, on ImageNet16 (NAS201), both Kendall’s tau and Recall@100 show steady improvements (from 0.55 to 0.63 and 13.67 to over 21, respectively). Overall, ConvNP demonstrates good data efficiency, achieving competitive performance with limited samples while still benefiting from increased training data, though with diminishing returns as performance gradually saturates.

\begin{table*}[h]
\centering
\caption{The performance of the ConvNP predictor under various settings on NAS101.} \label{Tb:nas101-ablation}
\footnotesize
\begin{tabular}{lcccccc}
\toprule
& MSELoss & RankingLoss &MLLoss &ML+Ranking  &NoSplit & NoSynthetic \\
\hline
Validation &	$94.33\pm0.20$ 	&	$93.97\pm0.46$  &  $94.44\pm0.19$ 	&    $94.69\pm0.20$   &   $94.46\pm0.85$     &	$94.27\pm0.21$ 			\\	
Test	   &    $93.75\pm0.21$	&   $93.44\pm0.44$  &  $93.85\pm0.19$   &    $94.04\pm0.13$	  &   $93.82\pm0.81$     &	$93.71\pm0.21$ 		\\
Ktau       &	$0.29$	        &	0.01 	        &  $0.46$           &	 $0.57$	          &   $0.54$             &	$0.28$ 	\\
R@100      &	$0.61$	        &	0.02         	&  $1.09$           &	 $5.75$ 	      &   $3.13$             &  $0.43$   \\
\bottomrule
\end{tabular}
\end{table*}

\subsection{Ablation studies}
Finally, we evaluate the performance of ConvNP under various settings. In each case, we vary only one component while keeping all other configurations fixed. The first four columns of Table~\ref{Tb:nas101-ablation} report results using MSE (Eq.~\ref{loss:mse}), Ranking (Eq.~\ref{loss:rank}), ML (Eq.~\ref{loss:nll}), and ML+Ranking losses on NAS101. Both MSE and Ranking alone perform poorly, with substantially lower accuracy and ranking quality compared to their combinations. In particular, the Ranking loss leads to extremely poor Kendall’s tau (0.01) and Recall@100 (0.02), highlighting its limitations. The ML loss achieves moderate performance (test accuracy 93.85\%), outperforming MSE but remaining inferior to combined objectives. ML+Ranking improves Recall@100 (5.75), although its validation and test accuracies are slightly lower than MSE+Ranking. We also observe that ML+Ranking converges more slowly, requiring multiple epochs, whereas MSE+Ranking achieves strong performance within a single epoch.

Columns 5 and 6 further examine the role of meta-learning components. Without context–target splitting (Column 5), where the entire task is treated as context, performance degrades noticeably (e.g., test accuracy drops to 93.82\%), indicating reduced generalization to unseen architectures. Similarly, removing synthetic task generation (Column 6) leads to a significant performance drop, suggesting overfitting to a single task and a lack of exposure to task variability. These results highlight the importance of both context–target splitting and synthetic task generation, confirming the effectiveness of the meta-learning formulation in ConvNP.

\section{Concluding remarks} \label{sec6}
In this paper, we propose a convolutional neural process (ConvNP)-based predictor for neural architecture search (NAS). Rather than learning a direct mapping from architectures to performance, we reformulate the problem as conditional function inference within a meta-learning framework. By constructing context–target splits over a distribution of synthetic tasks, ConvNP learns to generalize and predict the performance of unseen architectures. We evaluate our approach on NAS-Bench-101 and NAS-Bench-201, where it achieves competitive or superior performance compared to state-of-the-art predictors. Notably, our results reveal that strong global ranking metrics do not necessarily lead to better architecture selection. Instead, selection-oriented metrics, such as top-K performance, provide a more faithful assessment of practical effectiveness in NAS. We hope this perspective inspires future work on probabilistic and meta-learning-based predictors that better align with the practical objectives of neural architecture search.

\paragraph{Limitations}
While ConvNP achieves competitive performance, several limitations remain. First, its global ranking performance (e.g., Kendall’s tau) is slightly below specialized ranking-based predictors such as CARL, indicating room for improvement in fine-grained ordering. Second, the method relies on synthetic task generation and context–target splitting, and its performance may be sensitive to the design of these tasks. Third, compared to simple regressors, ConvNP introduces additional computational overhead due to its probabilistic and meta-learning structure. Finally, our evaluation is limited to NAS-Bench-101 and NAS-Bench-201, and further validation on larger or more diverse search spaces would strengthen the generality of our findings.

{\small 

\setcitestyle{numbers}
\bibliographystyle{plainnat}
\bibliography{nas.bib}
}

\end{document}